\documentclass{article} 
\usepackage{iclr2027_conference,times}

\usepackage{amsmath,amsfonts,bm}

\def\eqref#1{equation~\ref{#1}}

\def\1{\bm{1}}

\DeclareMathAlphabet{\mathsfit}{\encodingdefault}{\sfdefault}{m}{sl}
\SetMathAlphabet{\mathsfit}{bold}{\encodingdefault}{\sfdefault}{bx}{n}

\usepackage{hyperref}
\usepackage{url}

\usepackage{graphicx}

\usepackage{minitoc}
\usepackage{amsmath}
\usepackage{amssymb}
\usepackage{mathtools}
\usepackage{amsthm}
\usepackage{multirow}
\usepackage{xcolor}
\usepackage{colortbl}
\usepackage{booktabs}

\usepackage{algorithm}
\usepackage{algorithmicx}
\usepackage{algpseudocode}
\usepackage{verbatim}
\usepackage{subcaption}
\usepackage{fancyvrb}
\usepackage{fvextra}
\usepackage{xurl}
\usepackage{times}
\usepackage{helvet}
\usepackage{courier}
\usepackage[most]{tcolorbox}
\usepackage{wrapfig}
\usepackage{longtable}
\usepackage{afterpage}
\usepackage{subcaption}
\usepackage{algorithm}
\usepackage{algpseudocode}
\usepackage{array}
\usepackage{pifont}
\usepackage{CJKutf8}

\definecolor{GREEN}{HTML}{62b197}
\definecolor{RED}{HTML}{e18e6d}

\DeclareUnicodeCharacter{FF5C}{\textbar}

\definecolor{DarkGreen}{RGB}{0,120,0}
\definecolor{DarkRed}{RGB}{180,40,40}
\newcommand{\cmark}{\textcolor{DarkGreen}{\ding{51}}}
\newcommand{\xmark}{\textcolor{DarkRed}{\ding{55}}}

\newcommand{\ms}[2]{#1{{\scriptsize#2}}}

\newcommand{\spieval}{\textsc{SPIEval}}

\title{\spieval{}: Evaluating Large Language Models as Mobile Assistants over Scattered Personal Information}

\author{Junjie Ye$^{1,2}$, Zhuohui Sheng$^{1}$, Shaofan Liu$^{1}$, Yulun Zhu$^{1}$, Wenjie Fu$^{1}$,\\
\textbf{Dingwei Zhu$^{1}$, Ming Zhang$^{1}$, Yujiong Shen$^{1}$, Weichao Wang$^{2}$, Xin Zhao$^{2}$,}\\
\textbf{Shihan Dou$^{1}$, Tao Gui$^{1}$, Qi Zhang$^{1}$, Xuanjing Huang$^{1}$, Pluto Zhou$^{2}$}\\
\\
$^1$Fudan University, $^2$Tencent Hunyuan Team\\
\\
\texttt{jjye23@m.fudan.edu.cn}
}

\iclrfinalcopy 
\begin{document}

\maketitle

\begin{abstract}
Large language models (LLMs) are increasingly deployed as mobile assistants, where a key challenge is leveraging personal information scattered across multiple applications (apps) to complete user instructions. However, due to the lack of dedicated benchmarks, their capabilities remain poorly understood. To address this gap, we introduce \spieval{}, a human-curated benchmark grounded in five cognitive capabilities (i.e., reasoning, disambiguation, integration, preference inference, and multi-intent decomposition). \spieval{} comprises 250 tasks spanning 4,335 personal records distributed across 10 apps and supports multi-turn interaction through 21 tools. Analysis shows that the benchmark exhibits diverse scenarios, challenging tasks, scattered information, controllable environments, and verifiable outcomes. We evaluate nine representative LLMs and find substantial room for improvement. The best-performing model, GPT-5.5 (xhigh), achieves only 57.3\% accuracy, while the weakest achieves just 16.4\%. Further analysis reveals that 79\% of failures stem from inaccurate information localization, as LLMs often commit to plausible but incorrect information instead of continuing retrieval for verification. We also find that fewer than 2\% of retrieval actions employ advanced search methods and observe substantial variation in search efficiency across models. These findings expose fundamental limitations of current LLM-based mobile assistants and motivate future research in this direction. Data and code are available at~\url{https://huggingface.co/datasets/Junjie-Ye/SPIEval}.
\end{abstract}

\section{Introduction}
\begin{figure}[!t]
    \centering
    \includegraphics[width=\linewidth]{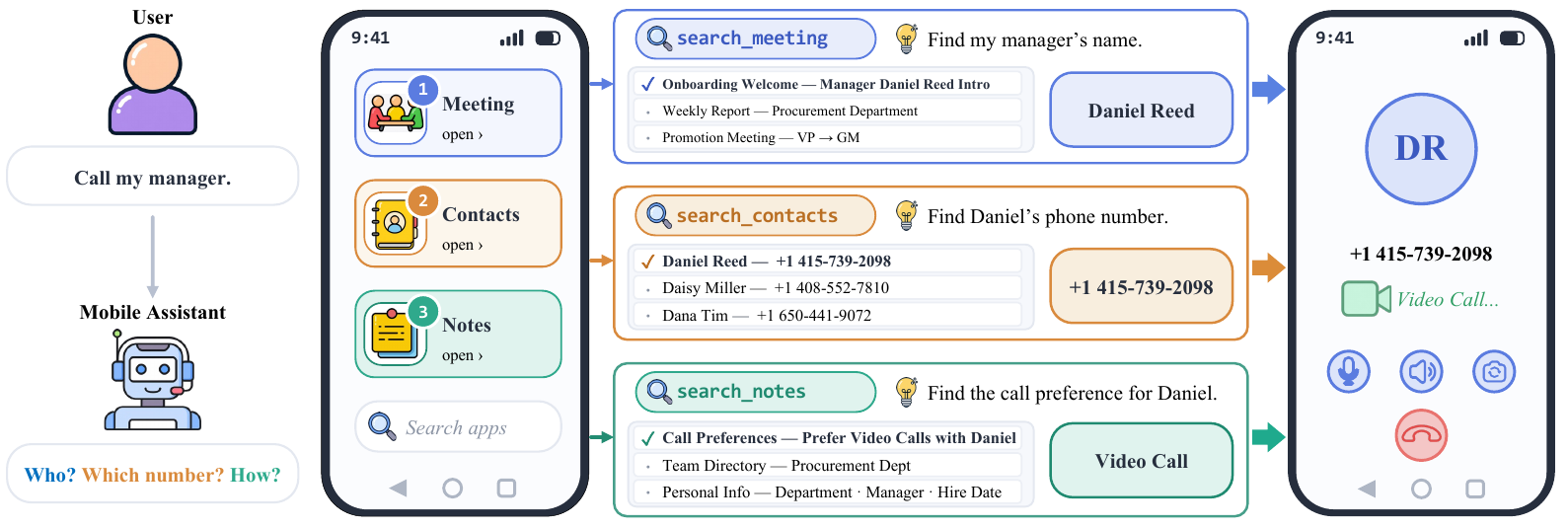}
    \caption{An example of a mobile assistant completing a user instruction by leveraging personal information scattered across multiple apps. To execute the instruction ``Call my manager,'' the assistant must proactively identify the manager, retrieve the corresponding phone number, and determine the preferred calling method from different apps before placing the call.}
    \vspace{-4mm}
    \label{fig:example}
\end{figure}

Due to their excellent instruction-following~\citep{analy-ye, IF-survey} and tool-use capabilities~\citep{tool-learning, tool-survey}, large language models (LLMs)~\citep{Claude-Opus-4.8, Gemini-3.1-pro, GPT5.5} are increasingly employed as mobile assistants~\citep{OS-survey, App-survey}. In this setting, they must leverage personal information scattered across multiple applications (apps) to complete users' brief and underspecified instructions. As shown in Figure~\ref{fig:example}, the instruction ``\textit{Call my manager}'' specifies neither the manager's identity nor the preferred calling method. To fulfill this request, the assistant must proactively identify the manager from meeting records, retrieve the corresponding phone number from the contacts app, and initiate a video call based on the user's preferences recorded in the notes app.

Significant research efforts have been devoted to evaluating LLM-based mobile assistants~\citep{AutoGLM, OSWorld, AndroidLab}. Some studies focus on app-operation capabilities. For instance, \citet{AppWorld} develop a simulated environment spanning nine apps to evaluate whether models can correctly orchestrate API calls. Other studies investigate the safety risks associated with app operation~\citep{SAPA}, while more recent efforts examine whether models can leverage personal data to generate personalized responses~\citep{HiCUPID}.

Despite targeting mobile assistants, these benchmarks primarily evaluate tool use and task execution under settings where the required information is explicitly provided or directly accessible. Specifically, user queries often provide the necessary parameters explicitly~\citep{AppWorld, OSWorld}, requiring models to map these parameters to appropriate API calls. Even when certain inputs are not specified in the initial instruction, they are typically returned directly by previous tool calls~\citep{ToolHop, AppAgent}. While personalization-oriented benchmarks introduce personal data, access to such information is generally limited to retrieving information from individual documents rather than locating information across multiple apps~\citep{PersonaBench}. As a result, these benchmarks place limited emphasis on the challenge of scattered personal information, leaving the effectiveness of LLM-based mobile assistants in this setting largely unexplored.

To address this gap, we propose \spieval{}, a human-curated benchmark for evaluating mobile assistants in scenarios with scattered personal information. The benchmark comprises 250 tasks covering five cognitive capabilities essential to this setting (i.e., reasoning, disambiguation, integration, preference inference, and multi-intent decomposition). These tasks are grounded in 4,335 records spanning 10 apps. To support multi-turn interactions, \spieval{} provides 21 tools, including 11 retrieval tools and 10 execution tools. Detailed analysis shows that \spieval{} features diverse scenarios, challenging tasks, scattered information, controllable environments, and verifiable outcomes, making it a rigorous benchmark for evaluating mobile assistants.

We conduct a comprehensive evaluation of nine representative LLMs. The results reveal substantial variation across models and reasoning efforts, indicating considerable room for improvement. Even the strongest model, GPT-5.5 (xhigh), achieves only 57.3\% accuracy, while the weakest achieves just 16.4\%. Further analysis shows that 79\% of failures arise from inaccurate localization of personal information, as LLMs tend to commit to plausible but incorrect information rather than continue retrieving for verification. We also find that fewer than 2\% of retrieval actions employ advanced search methods and observe substantial differences in search efficiency across models. These findings suggest that information localization constitutes the primary bottleneck for current mobile assistants and point to promising directions for future research.


\section{Related Work}
\textbf{LLM-Based Mobile Assistants~~}
Mobile assistants represent an important application of LLMs in personalized settings and have attracted widespread attention~\citep{OS-survey, App-survey}. Early efforts demonstrate that LLMs can automate multi-step mobile tasks by leveraging commonsense knowledge for action planning~\citep{AutoDroid}. Subsequent work extends these capabilities to real-world environments, enabling end-to-end task completion from natural language instructions~\citep{LLMPA}. More recent systems improve cross-app coordination through multi-agent architectures~\citep{MobileAgentV2, Fairy}, while self-evolving frameworks enable assistants to accumulate experience and continuously improve through interaction~\citep{AutoGLM, MobileAgentE}. These advances have also driven commercial adoption, with products such as Apple Intelligence\footnote{\url{https://www.apple.com/apple-intelligence/}}, Doubao Mobile Assistant\footnote{\url{https://o.doubao.com/}}, and Honor YOYO Agent\footnote{\url{https://www.honor.com/cn/magic-os/}} bringing intelligent task automation to hundreds of millions of users. As these systems become widely deployed, evaluating their capability to handle real-world personalized tasks becomes increasingly important. In this paper, we focus on the challenge of scattered personal information, where the information required to fulfill a user request is distributed across multiple apps.

\textbf{Evaluation of Mobile Assistants~~}
A growing body of work proposes benchmarks for evaluating LLM-based mobile assistants in app-centric settings. Early benchmarks focus on multi-step task execution through tool calls, evaluating whether agents can plan and orchestrate operations within simulated environments~\citep{AppWorld, Spa-Bench}. Subsequent efforts expand the evaluation scope along several dimensions, including dynamic conditions with asynchronous events and temporal constraints~\citep{Gaia2} as well as safety awareness during app operations~\citep{SAPA}. More recently, personalization-oriented benchmarks evaluate models' capabilities to leverage personal data for generating tailored responses~\citep{HiCUPID, PersonaBench}. However, these benchmarks either provide all necessary information directly in the user instructions or restrict access to personal information to simple document retrieval. As a result, they do not capture a common real-world setting in which fulfilling a user request requires locating information distributed across multiple apps. To address this gap, we introduce \spieval{}, a benchmark that evaluates mobile assistants under this challenging setting of scattered personal information.

\section{\spieval{}}

\begin{figure}[!t]
    \centering
    \includegraphics[width=\linewidth]{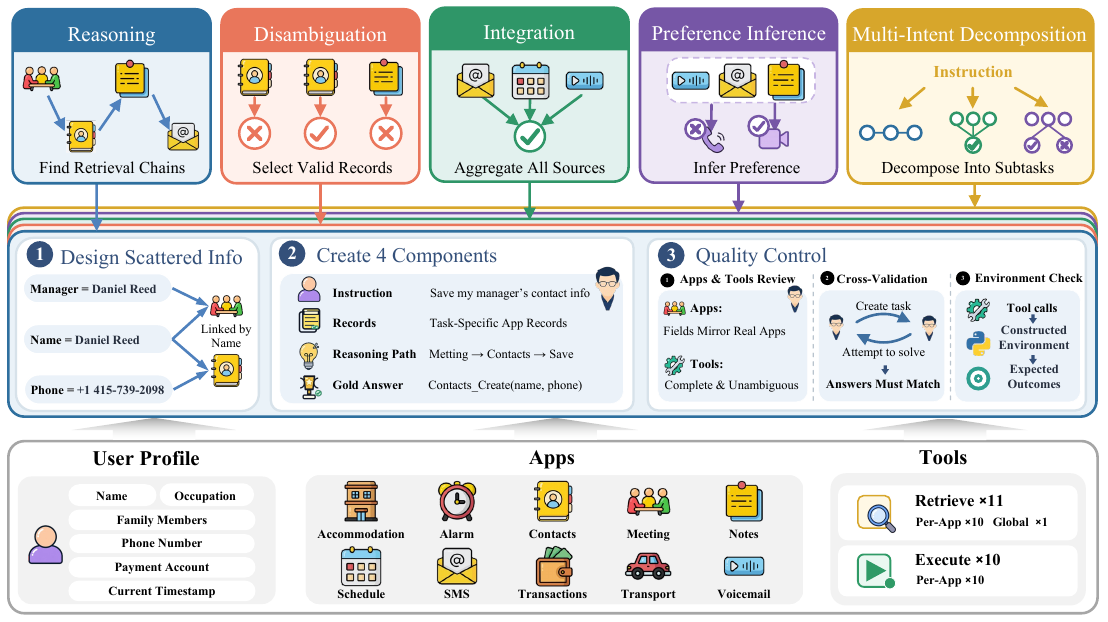}
    \caption{Framework of \spieval{}. \textbf{Bottom:} \spieval{} comprises 10 commonly used apps, together with 10 execution tools and 11 retrieval tools for accessing records across these apps. All records are associated with a unified user profile. \textbf{Middle:} Each task is designed to evaluate a specific cognitive capability. To ensure quality, the user instruction, personal records, reasoning process, and gold answer are manually constructed and independently verified by at least two annotators. \textbf{Top:} \spieval{} evaluates five cognitive capabilities essential for handling scattered personal information.}
    \vspace{-4mm}
    \label{fig:SPIEval}
\end{figure}


\subsection{Task Formulation}
\label{sec:formulation}
Given a natural-language user instruction $q$ and a set of apps $\mathcal{A} = \{a_1, a_2, \ldots, a_K\}$, where each app $a_k$ contains a set of structured personal records $\mathcal{R}_k$, the objective of a mobile assistant is to generate a sequence of tool calls that fulfills the instruction. In contrast to fully specified instructions, $q$ does not explicitly provide all information required to complete the task. The model is equipped with a set of tools $\mathcal{T} = \mathcal{T}_{\text{retrieve}} \cup \mathcal{T}_{\text{exec}}$ and must proactively formulate search queries to retrieve relevant records, reason over the retrieved information to infer the required parameters, and invoke execution tools with the inferred arguments. Formally, the model generates a trajectory $\tau = (c_1, r_1, c_2, r_2, \ldots, c_n, r_n)$, where each $c_i$ denotes a tool call and $r_i$ the corresponding tool feedback. The trajectory terminates with one or more execution calls.

\subsection{Cognitive Capabilities}
\label{sec:capabilities}

Figure~\ref{fig:SPIEval} summarizes five cognitive capabilities required for handling scattered personal information in mobile assistant settings.

\textbf{Reasoning~~}
Multi-hop reasoning, a capability extensively studied in other domains~\citep{HotpotQA}, is also vital for mobile assistants, where fulfilling a request typically requires a series of interdependent retrieval steps. The instruction ``\textit{Save my manager's contact information}'' can illustrate this dependency, since resolving the manager's identity enables retrieval of the corresponding phone number, which in turn supports the final save action.

\textbf{Disambiguation~~}
Disambiguation is a well-known challenge in information retrieval~\citep{Disambiguation}, and becomes even more difficult in mobile assistant settings due to ambiguous and evolving personal data. A possible case involves saving a contact's phone number when multiple numbers are associated with the same individual across different records or time periods. Completing this task requires identifying the currently valid number by leveraging contextual cues together with record-specific information.

\textbf{Integration~~}
In contrast to disambiguation, integration requires aggregating information from all relevant sources~\citep{FanOutQA}. Unlike the sequential dependency chains characteristic of multi-hop reasoning, these sources are often independent of one another. The instruction ``\textit{Save all suppliers from this week's trip}'' can exemplify this challenge, as it involves collecting supplier names and phone numbers that may be distributed across SMS messages, notes, and voicemail records.

\textbf{Preference Inference~~}
Preference inference is particularly important in personalized scenarios, as users often have habitual preferences that are never explicitly stated~\citep{Personalization}. A competent assistant must infer such preferences. This challenge may arise when a user says ``\textit{Call my wife},'' where the model must not only locate the appropriate contact information but also recognize based on prior call history that the user may typically prefer video calls over voice calls.

\textbf{Multi-Intent Decomposition~~}
Multi-intent decomposition refers to the capability to decompose a complex instruction into multiple independent subtasks~\citep{Decomposition}. Each subtask involves one of the preceding cognitive capabilities. The instruction ``\textit{Call Dad and transfer this month's living expenses}'' can combine a phone call and a payment transaction, requiring the model to decompose the request and retrieve the corresponding parameters from different sources.

\subsection{Benchmark Construction}
\label{sec:construction}

We construct \spieval{} through four components, followed by a rigorous quality control process.

\textbf{User Profile Construction~~}
To capture realistic mobile assistant usage scenarios, we construct a unified user profile that establishes a consistent identity across all tasks, which is provided to the model as part of the system prompt.\footnote{The complete user profile and system prompt are provided in Appendix~\ref{sec:appendix_profile}.} The profile specifies a set of core attributes, including the user's occupation, family members, a personal phone number, a payment account, and the current timestamp. These shared attributes allow user instructions to contain natural references such as ``my dad'' or ``my department manager,'' which the assistant must resolve correctly. Furthermore, each task is grounded in a distinct set of app records that introduce task-specific entities and information, such as colleagues, clients, and social contacts, thereby enabling diverse scenarios.

\textbf{Application Construction~~}
To reflect the diverse sources of personal information, we simulate 10 mobile apps commonly found on personal devices, including \textit{Accommodation}, \textit{Alarm}, \textit{Contacts}, \textit{Meeting}, \textit{Notes}, \textit{Schedule}, \textit{SMS}, \textit{Transactions}, \textit{Transport}, and \textit{Voicemail}. Each app is abstracted from its real-world counterpart and defined by a structured schema with domain-specific fields, comprising 8.1 fields on average.\footnote{The detailed schema of each app is provided in Appendix~\ref{sec:appendix_schema}.} In particular, each schema distinguishes required fields (e.g., a contact's name and phone number) from optional fields (e.g., company or birthday). Consequently, records are often only partially populated, and complete information about an entity may need to be assembled from records distributed across different apps. In addition, related fields are shared across apps, allowing records from different sources to be linked through common attributes such as names, phone numbers, and account numbers.

\begin{table}[!t]
\centering
\caption{Distribution of 357 execution operations across the five cognitive capabilities and ten operation categories in \spieval{}.}
\label{tab:diversity}
\resizebox{\columnwidth}{!}{%
\begin{tabular}{lccccccccccc}
\toprule
\textbf{Capability} & \textbf{\shortstack{Book\\Accommodation}} & \textbf{\shortstack{Set\\Alarm}} & \textbf{\shortstack{Make\\Call}} & \textbf{\shortstack{Save\\Contact}} & \textbf{\shortstack{Create\\Meeting}} & \textbf{\shortstack{Send\\Message}} & \textbf{\shortstack{Create\\Note}} & \textbf{\shortstack{Make\\Payment}} & \textbf{\shortstack{Create\\Schedule}} & \textbf{\shortstack{Book\\Transport}} & \textbf{Total} \\
\midrule
Reasoning & 5 & 5 & 5 & 5 & 5 & 6 & 5 & 5 & 5 & 5 & 51 \\
Disambiguation & 5 & 5 & 5 & 5 & 5 & 5 & 5 & 5 & 5 & 5 & 50 \\
Integration & 10 & 12 & 5 & 17 & 5 & 7 & 11 & 5 & 12 & 13 & 97 \\
Preference Inference & 5 & 5 & 5 & 5 & 5 & 5 & 5 & 5 & 5 & 8 & 53 \\
Multi-Intent Decomposition & 9 & 10 & 7 & 8 & 9 & 11 & 10 & 14 & 12 & 16 & 106 \\
\midrule
\textbf{Total} & 34 & 37 & 27 & 40 & 29 & 34 & 36 & 34 & 39 & 47 & 357 \\
\bottomrule
\end{tabular}%
}
\vspace{-4mm}
\end{table}

\textbf{Tool Construction~~}
To enable multi-turn interaction and support fine-grained analysis of model behavior, we design 21 tools organized into two complementary categories. One category comprises 11 retrieval tools that mirror the information access mechanisms available on mobile devices. Specifically, each app is equipped with a dedicated retrieval tool, alongside a global one. The per-app tools provide three retrieval modes, which are substring matching, regular expression matching, and fuzzy matching. They also support field-specific targeting and case-sensitivity control for precise information access. In contrast, the global tool performs substring-based search across all apps and returns results annotated with their source apps. All retrieval results are returned in a paginated manner, reflecting real-world search interfaces and requiring models to actively request additional results when necessary. The other category comprises 10 execution tools, one for each app, which serve as the execution endpoints for task completion. Each execution tool is accompanied by detailed parameter specifications, including type annotations and distinctions between required and optional fields, enabling parameter-level evaluation of action correctness.\footnote{The complete schema of all tools is provided in Appendix~\ref{sec:appendix_tools}.}

\textbf{Task Construction~~}
To ensure that each task admits a unique and verifiable solution, we adopt a fully manual construction process. For each task, an annotator first selects one of the five cognitive capabilities and designs the underlying information structure, specifying how task-relevant information is distributed across apps and connected through shared attributes. Based on this structure, the annotator constructs four components, including a natural-language user instruction, a set of task-specific app records, a step-by-step reasoning annotation, and a gold answer consisting of the exact execution tool calls with all parameters. The records are populated independently for each task such that all information required for task completion is available, but only through the intended retrieval and reasoning process. Following this procedure, we construct 250 tasks, with 50 tasks for each cognitive capability, grounded in 4,335 records distributed across 10 apps.

\textbf{Quality Control~~}
The benchmark is constructed over a period of three months by six NLP researchers. To guarantee the reliability and consistency of the benchmark, we implement quality control throughout the entire construction process. During the design of apps and tools, each schema is reviewed by all annotators to ensure that the abstracted fields faithfully reflect real-world app functionality and that tool definitions are complete and unambiguous. For task construction, we employ a cross-validation protocol in which one researcher creates a task and another independently attempts to solve it without access to the gold answer. A task is accepted only if both researchers arrive at the same answer through the intended retrieval process; otherwise, it is revised to eliminate ambiguities or unintended solution paths. In addition, all gold answers are executed against the tool implementation to verify that the corresponding tool calls produce the expected outcomes. The benchmark therefore admits 100\% human performance by construction. All personal data used in the benchmark is entirely fictional and does not correspond to any real individual.

\subsection{Dataset Analysis}
\label{sec:data_analysis}

To demonstrate that \spieval{} provides a rigorous evaluation of mobile assistants operating over scattered personal information, we analyze the benchmark along five dimensions.\footnote{A systematic comparison between \spieval{} and existing benchmarks is provided in Appendix~\ref{sec:appendix_comparison}.}

\textbf{Diverse Scenarios~~}
Mobile assistants are expected to handle a wide range of requests, requiring them to retrieve records through diverse strategies. A representative benchmark should cover a broad spectrum of tasks to enable systematic evaluation. To this end, each task is designed by jointly considering its required cognitive capability and execution operation. As shown in Table~\ref{tab:diversity}, the dataset comprises 357 execution operations spanning 10 operation categories. Each operation category is combined with all five cognitive capabilities, so that all 50 capability-operation pairs are represented, each occurring at least five times. Moreover, even the least frequent operation category appears 27 times, ensuring that no operation is underrepresented.

\begin{figure}[!t]
  \centering
  \begin{minipage}{0.45\textwidth}
    \centering
    \includegraphics[width=\linewidth]{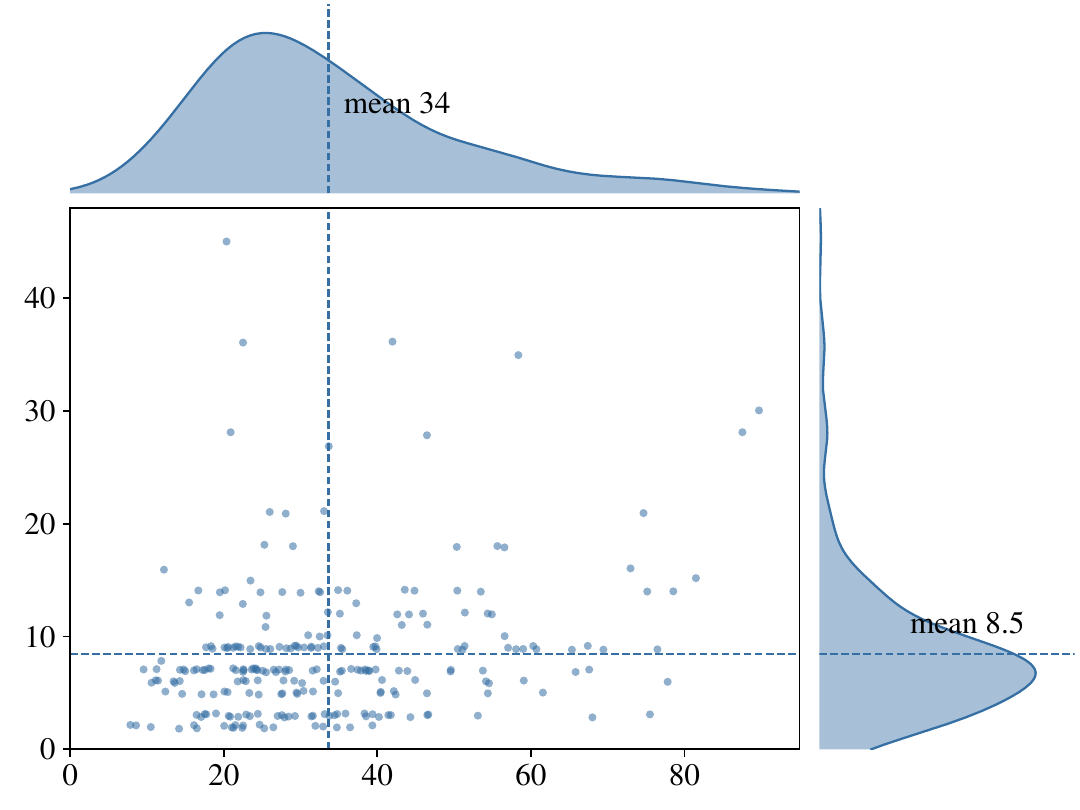}
    \caption{Relationship between instruction length and the total number of execution parameters required for each task, with marginal distributions shown alongside.}
    \label{fig:complexity}
  \end{minipage}
  \quad
  \begin{minipage}{0.45\textwidth}
    \centering
    \includegraphics[width=\linewidth]{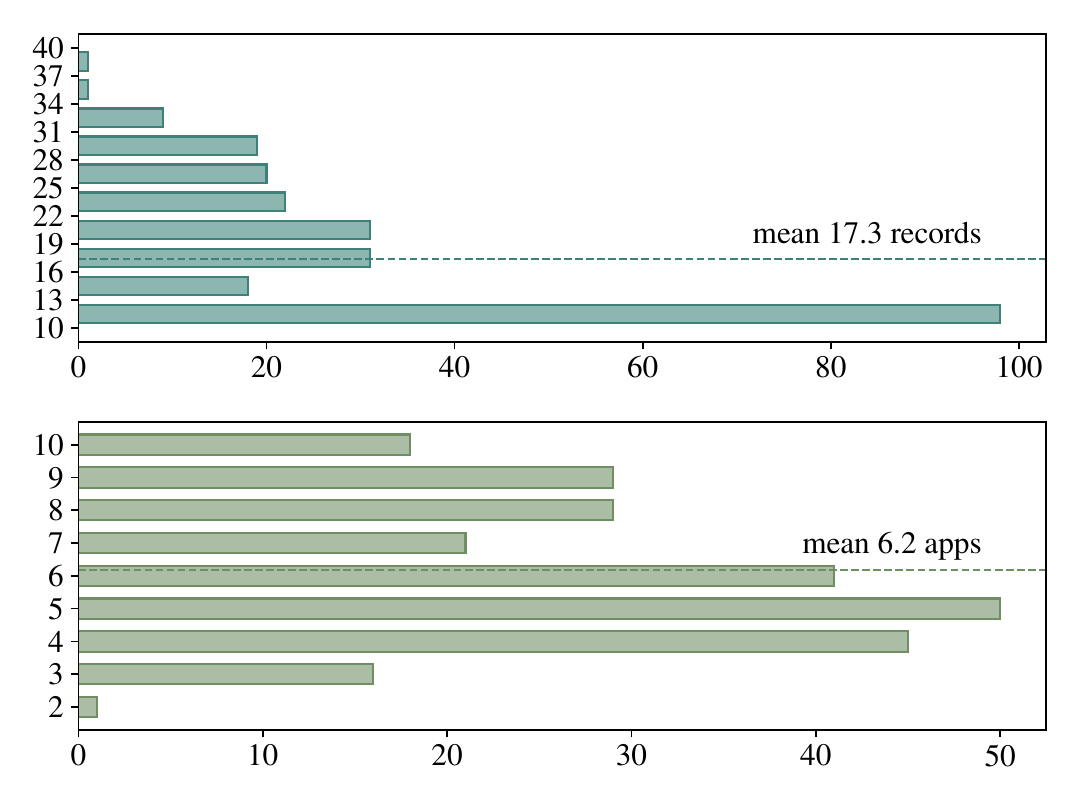}
    \caption{Distribution of task-relevant records across apps in \spieval{}. \textbf{Top:} Number of records per instruction. \textbf{Bottom:} Number of apps spanned by records.}
    \label{fig:scatter}
  \end{minipage}
  \vspace{-4mm}
\end{figure}

\textbf{Challenging Tasks~~}
Tasks for mobile assistants require models to interpret ambiguous user instructions, identify appropriate tools, infer the required parameters from the corresponding tool descriptions, and retrieve necessary information from personal data distributed across multiple apps, all of which are captured by \spieval{}. As shown in Figure~\ref{fig:complexity}, instructions in \spieval{} are highly concise, averaging only 34 characters, while the corresponding execution workflows require an average of 8.47 parameters. Notably, one instruction consists of only 20 characters yet requires 45 execution parameters. Moreover, user instructions explicitly provide none of the required parameters, leaving both tool selection and parameter inference largely implicit. Consequently, solving \spieval{} requires robust semantic understanding and effective information retrieval, making it representative of the challenging tasks encountered in the real world.

\textbf{Scattered Information~~}
Personal information on a mobile phone is stored in structured form across many apps, and task-relevant records are rarely confined to a single app. As a result, fulfilling an instruction requires retrieving and reconciling records from multiple apps. Figure~\ref{fig:scatter} quantifies this distribution. Each instruction in \spieval{} is associated with an average of 17.3 records spanning 6.2 of the 10 apps, with the most demanding instructions involving up to 39 records. This organization requires models to understand the relationships among apps, identify the relevant records, and integrate evidence from multiple sources to infer the necessary information.

\textbf{Controllable Environments~~}
A fair benchmark should attribute performance differences to the models themselves rather than inconsistencies in the runtime environment. Therefore, \spieval{} provides a fully controllable environment in which all records are preloaded and all retrieval and execution tools are implemented locally. Whenever a model invokes a tool, the environment validates the tool call against the tool specification and returns an informative error message if the input is invalid. Otherwise, the tool executes deterministically, ensuring that identical inputs always produce identical outputs. This design supports multi-turn interactions while ensuring reproducibility, enabling more reliable comparisons and targeted analyses of model behavior.

\textbf{Verifiable Outcomes~~}
A fundamental requirement of any benchmark is that its evaluation reflects model capability. To satisfy this requirement, \spieval{} avoids the LLM-as-a-judge paradigm, which is susceptible to model bias and output variability, and instead verifies model outputs against human-annotated gold answers. Each gold answer specifies the execution tools required for a task together with their parameter values, enabling automatic evaluation at the parameter level without subjective judgment. Certain parameters may take any valid value, such as the number of alarm repetitions, and are marked as unconstrained, in which case only their types are validated. For the remaining parameters, we annotate every acceptable value, and 7.0\% of them admit multiple valid values. For the 76 tasks that require multiple execution tools, the evaluation is invariant to tool invocation order. Together, these design choices provide a consistent, transparent, and reproducible evaluation protocol for comparing different models.

\begin{table}[!t]
\centering
\caption{Main results on \spieval{}. Each model is evaluated under its highest and lowest reasoning effort levels, indicated in parentheses. Human accuracy is 100\% for all categories by construction.}
\label{tab:main_results}
\resizebox{\columnwidth}{!}{%
\begin{tabular}{lcccccc}
\toprule
\multirow{2}{*}{\textbf{Model}} & \multirow{2}{*}{\textbf{Reasoning}} & \textbf{Disam-} & \multirow{2}{*}{\textbf{Integration}} & \textbf{Preference} & \textbf{Multi-Intent} & \multirow{2}{*}{\textbf{Overall}} \\
& & \textbf{biguation} & & \textbf{Inference} & \textbf{Decomposition} & \\
\midrule
GPT-5.5 (xhigh) & \ms{70.0}{4.3} & \ms{72.0}{0.0} & \ms{73.3}{0.9} & \ms{\textbf{38.7}}{2.5} & \ms{\textbf{32.7}}{2.5} & \ms{\textbf{57.3}}{1.2} \\
Gemini 3.1 Pro (high) & \ms{\textbf{74.0}}{4.3} & \ms{\textbf{73.3}}{0.9} & \ms{\textbf{74.7}}{1.9} & \ms{20.0}{4.9} & \ms{23.3}{1.9} & \ms{53.1}{2.2} \\
Claude Opus 4.8 (max) & \ms{68.7}{3.4} & \ms{58.7}{3.4} & \ms{67.3}{1.9} & \ms{34.7}{4.1} & \ms{32.0}{1.6} & \ms{52.3}{0.7} \\
DeepSeek-V4-Pro (max) & \ms{53.3}{0.9} & \ms{62.7}{4.1} & \ms{60.0}{4.3} & \ms{34.0}{2.8} & \ms{22.7}{1.9} & \ms{46.5}{0.4} \\
Gemini 3.1 Pro (low) & \ms{65.3}{1.9} & \ms{62.7}{5.0} & \ms{60.7}{6.2} & \ms{16.7}{4.1} & \ms{23.3}{3.8} & \ms{45.7}{2.2} \\
Claude Opus 4.8 (none) & \ms{48.7}{3.4} & \ms{40.7}{5.0} & \ms{46.0}{3.3} & \ms{26.7}{2.5} & \ms{19.3}{0.9} & \ms{36.3}{1.7} \\
Kimi K2.6 (high) & \ms{48.7}{1.9} & \ms{46.7}{6.6} & \ms{43.3}{0.9} & \ms{25.3}{2.5} & \ms{16.0}{1.6} & \ms{36.0}{2.3} \\
GLM-5.2 (max) & \ms{44.7}{5.0} & \ms{44.7}{1.9} & \ms{49.3}{0.9} & \ms{22.7}{5.2} & \ms{14.7}{3.4} & \ms{35.2}{1.4} \\
Hy3 (high) & \ms{44.0}{2.8} & \ms{41.3}{1.9} & \ms{46.0}{4.3} & \ms{26.0}{2.8} & \ms{14.0}{1.6} & \ms{34.3}{1.9} \\
DeepSeek-V4-Pro (none) & \ms{36.0}{2.8} & \ms{39.3}{1.9} & \ms{48.7}{5.0} & \ms{28.7}{1.9} & \ms{18.7}{0.9} & \ms{34.3}{1.4} \\
Qwen3.7-Plus (high) & \ms{44.0}{2.8} & \ms{38.7}{0.9} & \ms{46.7}{3.8} & \ms{23.3}{3.4} & \ms{14.7}{3.8} & \ms{33.5}{0.7} \\
Seed-2.1-Pro (high) & \ms{50.0}{4.9} & \ms{44.0}{4.3} & \ms{39.3}{2.5} & \ms{17.3}{0.9} & \ms{16.0}{2.8} & \ms{33.3}{1.0} \\
GLM-5.2 (none) & \ms{37.3}{2.5} & \ms{32.0}{4.3} & \ms{46.7}{2.5} & \ms{19.3}{1.9} & \ms{10.7}{2.5} & \ms{29.2}{1.1} \\
GPT-5.5 (none) & \ms{41.3}{5.2} & \ms{35.3}{3.4} & \ms{30.7}{4.7} & \ms{22.0}{2.8} & \ms{13.3}{3.8} & \ms{28.5}{1.1} \\
Seed-2.1-Pro (minimal) & \ms{34.7}{2.5} & \ms{32.7}{1.9} & \ms{27.3}{5.0} & \ms{18.7}{4.1} & \ms{9.3}{2.5} & \ms{24.5}{0.8} \\
Qwen3.7-Plus (none) & \ms{34.7}{3.4} & \ms{27.3}{2.5} & \ms{36.0}{1.6} & \ms{12.7}{1.9} & \ms{8.7}{0.9} & \ms{23.9}{0.5} \\
Hy3 (no\_think) & \ms{25.3}{2.5} & \ms{24.7}{1.9} & \ms{23.3}{10.5} & \ms{13.3}{5.0} & \ms{6.0}{1.6} & \ms{18.5}{1.6} \\
Kimi K2.6 (none) & \ms{17.3}{0.9} & \ms{26.0}{2.8} & \ms{19.3}{7.5} & \ms{11.3}{2.5} & \ms{8.0}{0.0} & \ms{16.4}{2.0} \\
\bottomrule
\end{tabular}%
}
\vspace{-4mm}
\end{table}

\section{Experimental Setup}
\label{sec:setup}
We describe the models, metrics, and implementation details to facilitate reproducibility.

\textbf{Models~~}
We evaluate nine representative state-of-the-art LLMs, including {Claude Opus 4.8}~\citep{Claude-Opus-4.8}, {DeepSeek-V4-Pro}~\citep{DeepSeek-V4}, {Gemini 3.1 Pro}~\citep{Gemini-3.1-pro}, {GLM-5.2}~\citep{GLM-5.2}, {GPT-5.5}~\citep{GPT5.5}, {Hy3}~\citep{Hy3}, {Kimi K2.6}~\citep{Kimi-K2.6}, {Qwen3.7-Plus}~\citep{qwen37plus}, and {Seed-2.1-Pro}~\citep{Seed2.1}.

\textbf{Metrics~~}
To objectively evaluate model performance, we adopt a binary accuracy metric. Because a model may adapt its behavior based on tool feedback, evaluating intermediate steps is neither necessary nor appropriate. Instead, we compare the model's final execution tool calls against the annotated gold answers. An outcome is considered correct only if the invoked execution tools and all of their parameter values exactly match one of the annotated gold answers.

\textbf{Implementation Details~~}
For each model, we evaluate performance under both the highest and lowest available reasoning effort levels. All other hyperparameters are kept at their default values to maximize each model's performance. Since mobile assistants are subject to latency constraints, we limit each task to a maximum of 50 interaction turns, balancing sufficient interaction with the environment against unnecessary computation. All retrieval results are returned in a paginated manner, with at most five results per page, requiring models to actively request additional pages when needed. To mitigate sampling variability, we conduct three independent runs for each experimental setting and report the mean and standard deviation across runs.

\section{Main Results}
\label{sec:main_results}

\begin{figure}[!t]
  \centering
  \begin{minipage}{0.45\textwidth}
    \centering
    \includegraphics[width=\linewidth]{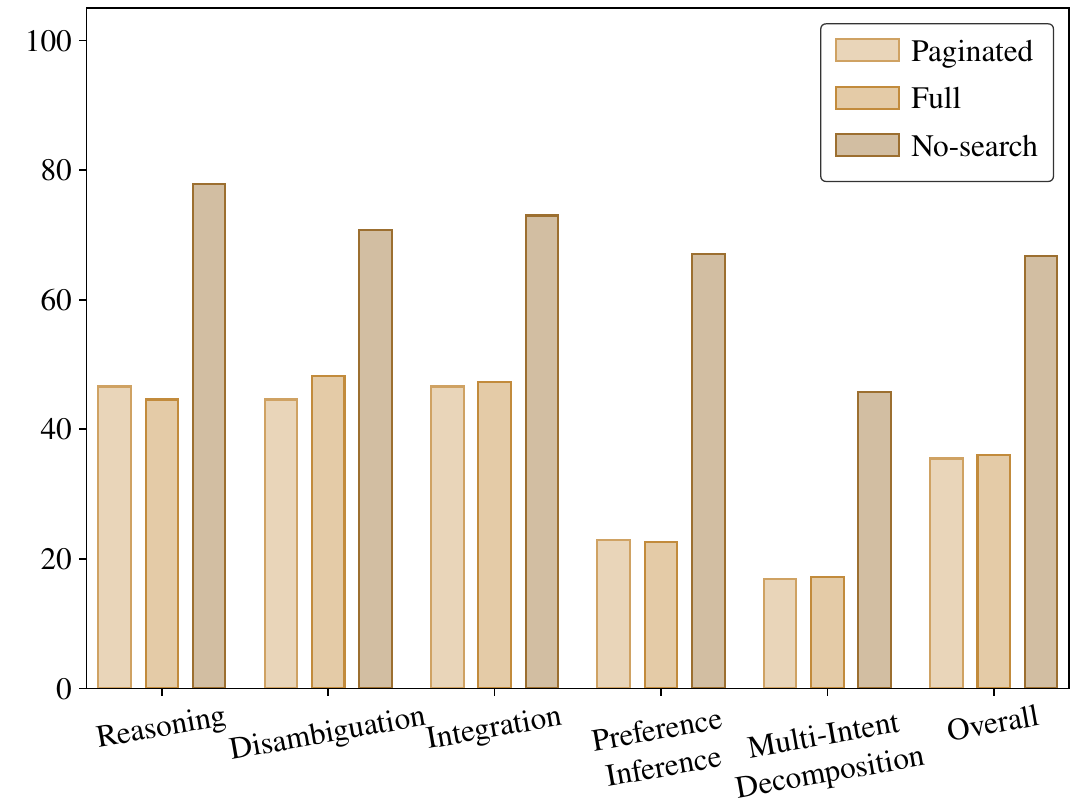}
    \caption{Accuracy under the standard paginated protocol and two controlled settings, averaged over all model configurations.}
    \label{fig:mode_comparison}
  \end{minipage}
  \quad
  \begin{minipage}{0.45\textwidth}
    \centering
    \includegraphics[width=\linewidth]{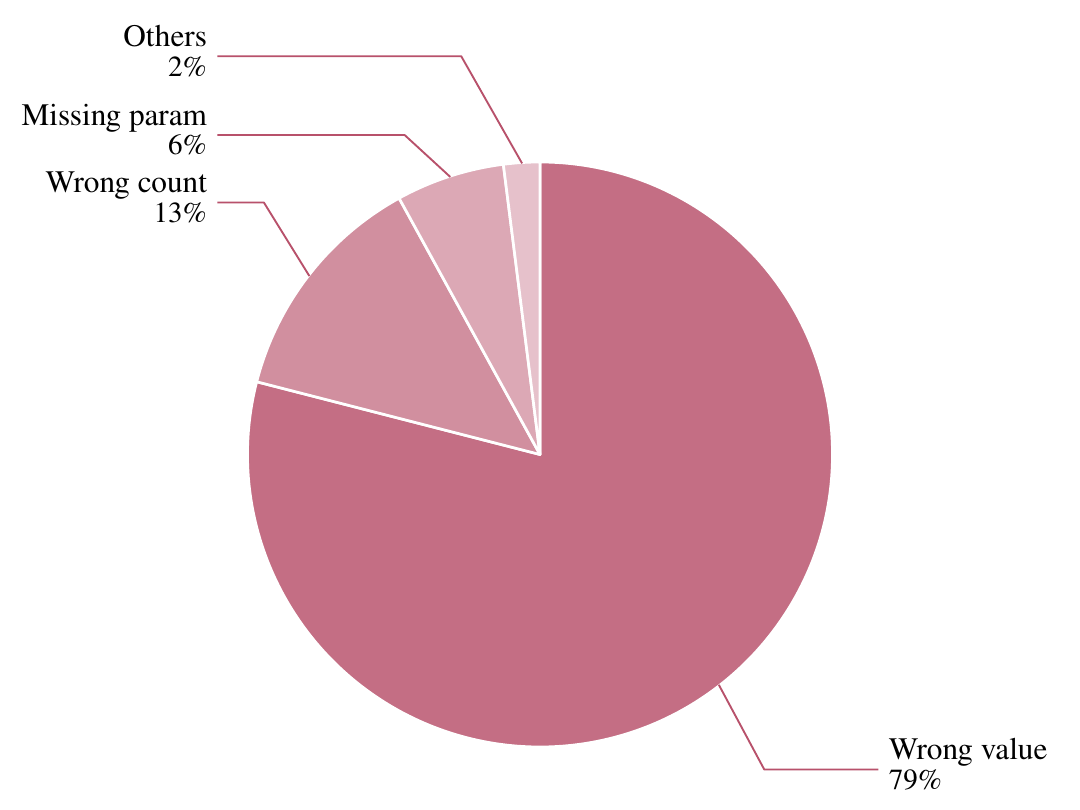}
    \caption{Distribution of error types from GPT-5.5 (xhigh), Gemini 3.1 Pro (high), and Claude Opus 4.8 (max).}
    \label{fig:error_dist}
  \end{minipage}
  \vspace{-4mm}
\end{figure}

Table~\ref{tab:main_results} summarizes the performance of LLMs, from which we draw the following observations.

\textbf{Current LLMs struggle with information localization, making \spieval{} a challenging benchmark.}
Even GPT-5.5 (xhigh), the strongest evaluated model, achieves an accuracy of only 57.3\%. Meanwhile, relatively weaker models, such as Kimi K2.6 (none), achieve only 16.4\% accuracy, rendering them largely impractical for this task. To better understand the source of this difficulty, we evaluate models under two additional settings. In the \emph{full} setting, retrieval tools return all matching records in a single response instead of presenting them page by page. In the \emph{no-search} setting, all task-relevant records are provided directly in the system prompt, eliminating retrieval altogether. As shown in Figure~\ref{fig:mode_comparison}, removing the retrieval process increases the average accuracy from 35.5\% to 66.8\%, whereas returning all matching records at once improves accuracy only marginally, from 35.5\% to 36.0\%. These results suggest that the primary bottleneck stems from the capability to formulate effective queries that correctly identify the target record. We further analyze the failure modes of the three strongest models. As shown in Figure~\ref{fig:error_dist}, 79\% of all failures arise from incorrect parameter values, whereas incorrect tool selection is relatively uncommon, and only 6\% of failures result from mandatory parameters being left unspecified. These findings suggest that LLMs often commit to plausible but incorrect information instead of continuing retrieval for verification. Such overconfident behavior poses a serious reliability risk for mobile assistants, where acting on incorrect personal information may be more harmful than declining to act.

\textbf{LLMs exhibit consistent performance differences across cognitive capabilities.}
LLMs achieve an average accuracy of around 46\% on reasoning, disambiguation, and integration, whereas their average performance on preference inference and multi-intent decomposition is only about half as high. This disparity reflects a fundamental difference between these capabilities. The former primarily requires identifying and combining information explicitly available in personal records, whereas the latter requires inferring information that is not explicitly stated. Figure~\ref{fig:mode_comparison} further distinguishes the sources of difficulty for preference inference and multi-intent decomposition. Under the \emph{no-search} setting, performance on preference inference reaches 67.0\%, representing a 44.1-point improvement over the standard setting. This result suggests that LLMs are capable of inferring user preferences once the relevant evidence is available, but struggle to proactively locate that evidence, a capability that is essential for mobile assistants. By contrast, even with all relevant records directly available, performance on multi-intent decomposition reaches only 45.7\%. This finding indicates that its primary bottleneck lies in decomposing complex requests into coordinated subtasks and planning their execution, another fundamental capability for mobile assistants.

\textbf{Increasing reasoning effort improves performance, although the gains vary substantially across models.}
Across all evaluated models, increasing the reasoning effort consistently improves performance, yielding an average gain of 13.8 points. However, these gains vary considerably across models, ranging from 28.8 points for GPT-5.5 to only 6.0 points for GLM-5.2. This variation suggests that the benefits of additional reasoning depend on a model's ability to translate extra deliberation into more effective actions. In \spieval{}, this capability is reflected in how models use the additional reasoning budget to plan retrieval strategies, verify intermediate results, and reformulate queries when the initial search fails to identify the desired records. Models with smaller improvements either already retrieve information efficiently with minimal reasoning, as exemplified by Gemini 3.1 Pro, or fail to convert additional reasoning into better decisions. These findings suggest that the ability to translate additional reasoning into more effective information localization and retrieval is a key advantage for LLMs intended to function as mobile assistants.

\section{Further Analysis}
\label{sec:analysis}

To better understand the factors underlying model performance, we conduct a more in-depth analysis and draw the following observations.

\begin{figure}[!t]
  \centering
  \begin{minipage}{0.45\textwidth}
    \centering
    \includegraphics[width=\linewidth]{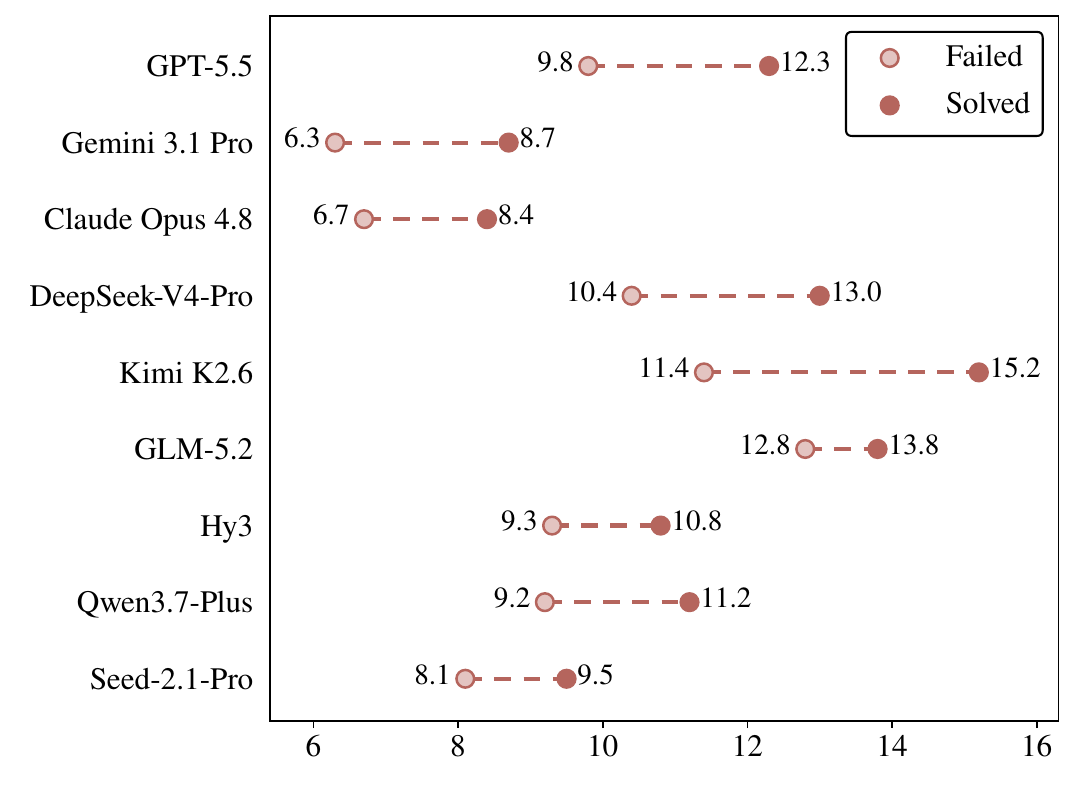}
    \caption{Average number of retrievals on solved versus failed tasks, reported for each model under its highest reasoning effort.}
    \label{fig:search_counts}
  \end{minipage}
  \quad
  \begin{minipage}{0.45\textwidth}
    \centering
    \includegraphics[width=\linewidth]{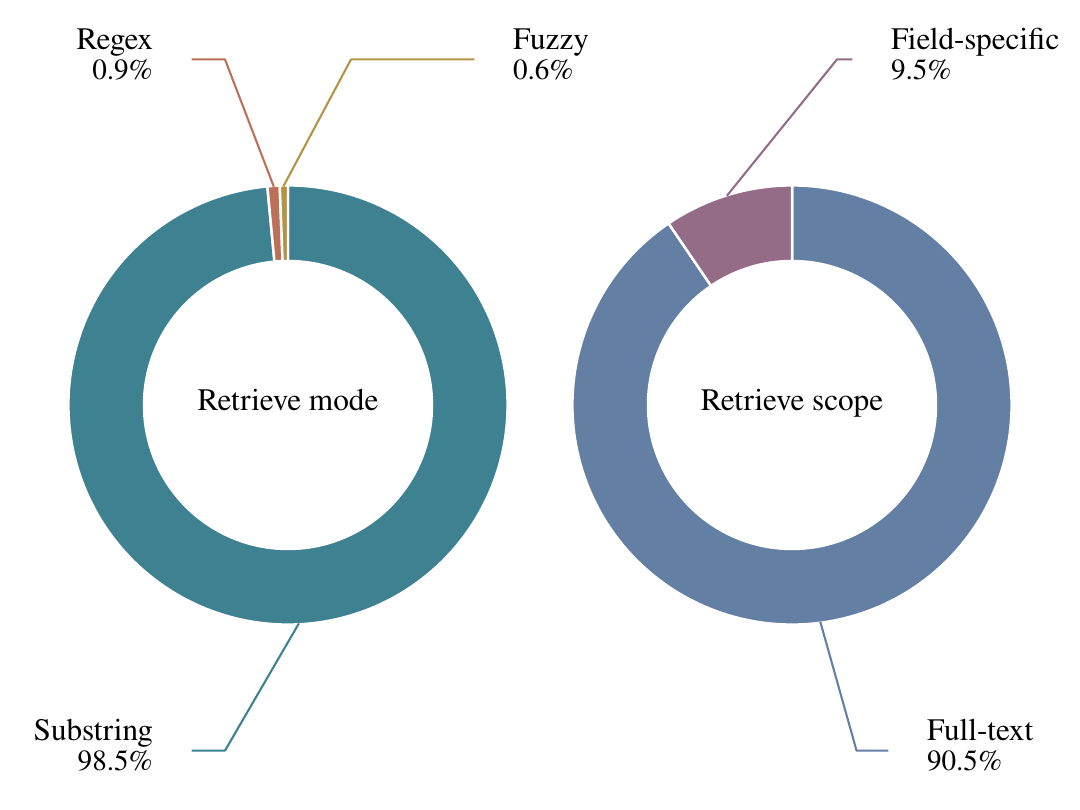}
    \caption{Distributions of retrieval configurations across all 126,279 retrieval calls issued by 18 model configurations.}
    \label{fig:search_methods}
  \end{minipage}
  \vspace{-4mm}
\end{figure}

\textbf{LLMs often fail because they commit to decisions too early.}
By comparing retrieval behavior on successful and failed tasks, we observe a counterintuitive pattern. As shown in Figure~\ref{fig:search_counts}, every evaluated model performs fewer retrievals on failed tasks than on successful ones, with the difference ranging from 1.0 retrieval for GLM-5.2 to 3.8 retrievals for Kimi K2.6. Combined with the findings in Figure~\ref{fig:error_dist}, this pattern suggests that models do not fail because they give up on difficult tasks. Instead, they often stop searching as soon as they encounter a seemingly plausible record and proceed with the requested action without performing additional retrievals to verify the information or distinguish it from competing candidates. These findings suggest that an important direction for future models is to develop their capability to determine whether the available evidence is sufficient and decide when additional retrieval is necessary.

\textbf{LLMs make little use of the advanced retrieval methods provided by the tools.}
A comprehensive analysis of all 126,279 retrieval calls issued across 18 model configurations, shown in Figure~\ref{fig:search_methods}, reveals that plain substring queries account for 98.5\% of all retrievals. By comparison, regular expressions and fuzzy matching together account for less than 2\%, while only 9.5\% of retrievals restrict the search to specific fields. One possible explanation is that the models inherit keyword-based retrieval strategies from general-purpose text retrieval rather than learning to exploit the structured retrieval capabilities exposed by the tools. This overwhelming reliance on basic substring matching has practical consequences. Records containing noisy values, lexical variations, or only partial matches often cannot be retrieved through exact substring matching alone. Since advanced methods are already exposed through the retrieval interface, enabling models to use them more effectively represents a promising direction for improving future performance.

\textbf{Different LLMs exhibit distinct strengths in reasoning and retrieval behavior.}
Gemini 3.1 Pro achieves accuracy above 73\% on reasoning, disambiguation, and integration, but its performance drops to 20\% on preference inference. GPT-5.5 exhibits the opposite trend, achieving the strongest performance on preference inference and multi-intent decomposition. Meanwhile, Gemini 3.1 Pro and Claude Opus 4.8 perform an average of 7.6 retrievals per task, whereas GPT-5.5 performs 11.2 retrievals per task on average. Despite using only about two-thirds as many retrievals, Gemini 3.1 Pro and Claude Opus 4.8 achieve performance comparable to GPT-5.5, suggesting that they rely on more targeted retrieval strategies. By contrast, GPT-5.5 achieves the highest overall accuracy through more comprehensive retrieval. These findings suggest that successful mobile assistants can adopt different strategies, balancing retrieval efficiency against comprehensive information gathering and downstream reasoning.

\section{Conclusion}
\label{sec:conclusion}

In this paper, we introduce \spieval{}, a benchmark for evaluating LLMs as mobile assistants that complete tasks by leveraging scattered personal information. The benchmark covers five cognitive capabilities and comprises 4,335 personal records collected from 10 commonly used apps. We evaluate nine representative LLMs and find that current models struggle in this setting, with information localization emerging as the primary bottleneck. Further analyses reveal systematic limitations in retrieval behavior, reasoning strategies, and capability utilization, providing a deeper understanding of current LLM-based mobile assistants.


\bibliography{iclr2027_conference}
\bibliographystyle{iclr2027_conference}

\clearpage

\appendix

\begin{CJK}{UTF8}{gbsn}

\section{User Profile and System Prompt}
\label{sec:appendix_profile}

As described in Section~\ref{sec:construction}, we construct a unified user profile that establishes a consistent identity across all tasks. The profile is provided to the model as part of the system prompt. Table~\ref{tab:user_profile} presents the user profile, and Figure~\ref{fig:system_prompt} shows the complete system prompt template.

\begin{table}[h]
\centering
\caption{Unified user profile provided as part of the system prompt for all tasks.}
\label{tab:user_profile}
\begin{tabular}{lll}
\toprule
\textbf{Attribute} & \textbf{Chinese} & \textbf{English} \\
\midrule
Name & 徐艺轩 & Yixuan Xu \\
\multirow{2}*{Occupation} & \multirow{2}*{上海光明坚果公司采购部普通员工} & Employee, Procurement Dept., \\
& & Shanghai Guangming Nut Co. \\
\multirow{2}*{Family Members} & \multirow{2}*{妻子谢玲、儿子徐鑫、女儿徐淼} & Wife: Ling Xie; Son: Xin Xu; \\
& & Daughter: Miao Xu \\
Phone Number & 15711227837 & 15711227837 \\
Payment Account & 6977889967897890098 & 6977889967897890098 \\
Current Timestamp & 2025-11-01 12:00:00 周六 & 2025-11-01 12:00:00 Saturday \\
\bottomrule
\end{tabular}
\end{table}

\begin{center}
\begin{tcolorbox}[
    colback=white,
    colframe=gray!70!black,
    title=System Prompt Template,
    coltitle=white,
    fonttitle=\bfseries,
    center title,
    rounded corners,
    boxrule=0.6mm,
    width=\linewidth,
    breakable,
    enhanced,
    left=6pt,
    right=6pt,
    top=4pt,
    bottom=4pt
]
\textbf{Chinese:}

\begin{Verbatim}[breaklines]
你是一个智能手机助手。用户会给你一些指令，你需要通过查询手机中各个应用的记录来收集必要信息，然后调用相应的工具完成用户的请求。

{user_profile}

你可以使用以下两类工具：
1. **搜索工具**：用于在手机各应用中检索记录，包括{len(app_tools)}个应用内搜索工具（{app_list}）和1个全局搜索工具（{global_tool}）。
2. **执行工具**：用于执行具体操作，包括{exec_list}。

工作流程：
1. 分析用户意图，判断需要哪些信息
2. 调用搜索工具查找相关记录，可以多次搜索、组合使用不同的搜索工具
3. 根据搜索结果，调用执行工具完成用户请求
4. 如果用户的请求包含多个意图，请并行调用多个执行工具同时完成

注意事项：
- 完成用户请求所需的所有信息都可以从手机记录中检索得到，请充分搜索，不要向用户反问
- 搜索结果会按页返回，如果提示还有更多结果，可通过page参数翻页查看
- 如果一次搜索未找到所需信息，尝试换个关键词或换个应用搜索
- 多个搜索之间如果相互独立，可以并行调用
- 最终的执行工具调用应包含尽可能完整的参数信息
- 当所有操作完成后，请回复“[DONE]需求已完成，还有什么可以帮您？”


\end{Verbatim}

\textbf{English:}

\begin{Verbatim}[breaklines]
You are a mobile assistant. The user will give you instructions. You need to retrieve the necessary information from records across different phone applications before invoking the appropriate tools to fulfill the user's request.

{user_profile}

You have access to two categories of tools:
1. **Retrieval tools**: used to retrieve records from phone applications, including {len(app_tools)} app-specific retrieval tools ({app_list}) and one global retrieval tool ({global_tool}).
2. **Execution tools**: used to perform specific operations, including {exec_list}.

Workflow: 
1. Analyze the user's intent and determine what information is needed
2. Call search tools to find relevant records---you may issue multiple retrieval queries and combine different retrieval tools 3. Based on the search results, call execution tools to fulfill the user's request
4. If the user's request contains multiple intents, invoke multiple execution tools in parallel

Notes: 
- All information needed to fulfill the user's request can be retrieved from phone records. Search thoroughly instead of asking the user for clarification. 
- Search results are returned in pages; if more results are available, use the page parameter to view additional pages. 
- If a retrieval does not return the required information, try different keywords or a different application. 
- Independent retrievals may be performed in parallel. 
- The final execution tool calls should include all available parameter values whenever possible.
- Once all tasks are complete, please reply with "[DONE] Request completed. Is there anything else I can help you with?"

\end{Verbatim}
\end{tcolorbox}

\captionof{figure}{System prompt template used for experiments.}
\label{fig:system_prompt}
\end{center}

\clearpage

\section{Application Schemas}
\label{sec:appendix_schema}

As described in Section~\ref{sec:construction}, we construct 10 simulated apps. Each app is abstracted from its real-world counterpart and represented by a structured schema containing domain-specific fields, with an average of 8.1 fields per app. Each schema further distinguishes between required and optional fields. Table~\ref{tab:app_schema} provides the complete schema of the 10 simulated apps in \spieval{}.

\begin{table}[h]
\centering
\caption{Schemas of the 10 simulated apps in \spieval{}, including required and optional fields.}
\label{tab:app_schema}
{
\begin{tabular}{m{0.16\linewidth}m{0.37\linewidth}m{0.37\linewidth}}
\toprule
\textbf{App} & \textbf{Required Fields} & \textbf{Optional Fields} \\
\midrule
\textit{Accommodation} 
& Check-in Date, Check-out Date, Contact, Contact Number, Cost, Hotel, Quantity, Room Type, Status
& -- \\

\textit{Alarm} 
& Alarm Name, Repeat, Repeat Count, Ring Duration, Ring Interval, Status, Time
& -- \\

\textit{Contacts} 
& Name, Phone Number
& Address, Birthday, Company, Email, ID Number, Note, Relationship \\

\textit{Meeting} 
& End Date, End Time, Meeting ID, Participants, Reminder, Repeat, Start Date, Start Time, Title
& Location, Meeting Minutes, Status \\

\textit{Notes} 
& Content
& Creation Date, Creation Time, Title \\

\textit{Schedule} 
& Date, End Time, Reminder, Repeat, Start Time, Title
& Location \\

\textit{SMS} 
& Content, Date, Recipient Phone, Status, Time
& Attachment, Recipient Name, Sender Name, Sender Phone \\

\textit{Transactions} 
& Amount, Date, Item, Payee Account, Payer, Payer Account, Time
& Note, Payee \\

\textit{Transport} 
& Contact, Contact Number, Date, Departure Time, Destination, Estimated Arrival Time, Estimated Cost, Mode, Origin, Status
& -- \\

\textit{Voicemail} 
& Caller Number, Date, Message Content, Time
& Caller Name \\
\bottomrule
\end{tabular}
}
\end{table}

\clearpage

\section{Tool Schemas}
\label{sec:appendix_tools}

As described in Section~\ref{sec:construction}, we construct 21 tools, including 11 retrieval tools and 10 execution tools. Each app is associated with one app-specific retrieval tool and one execution tool, while an additional global retrieval tool enables cross-app information access. To provide a detailed description of these tools, we present their specification documents in Figure~\ref{fig:retrieval_tools} and Figure~\ref{fig:execution_tools}, respectively.

\begin{center}
\begin{tcolorbox}[
    colback=white,
    colframe=gray!70!black,
    title=Specification Documents for all Retrieval Tools,
    coltitle=white,
    fonttitle=\bfseries,
    center title,
    rounded corners,
    boxrule=0.6mm,
    width=\linewidth,
    breakable,
    enhanced,
    left=6pt,
    right=6pt,
    top=4pt,
    bottom=4pt
]

\begin{Verbatim}[breaklines]
[
    {
        "type": "function",
        "function": {
            "name": "search_accommodation",
            "description": "在住宿记录中搜索。",
            "parameters": {
                "type": "object",
                "properties": {
                    "pattern": {
                        "type": "string",
                        "description": "搜索内容。"
                    },
                    "field": {
                        "type": "string",
                        "description": "指定搜索的字段名。可选字段：'住宿地点'、'入住日期'、'退宿日期'、'房型'、'数量'、'联系人'、'联系方式'、'费用'、'状态'。不指定则搜索所有字段。"
                    },
                    "mode": {
                        "type": "string",
                        "description": "匹配模式。fixed：子串包含匹配（默认）；regex：正则表达式匹配；fuzzy：模糊相似度匹配。",
                        "enum": [
                            "fixed",
                            "regex",
                            "fuzzy"
                        ]
                    },
                    "ignore_case": {
                        "type": "boolean",
                        "description": "是否忽略大小写，默认为true。"
                    },
                    "page": {
                        "type": "integer",
                        "description": "返回第几页结果，每页最多5条。默认为1。"
                    }
                },
                "required": [
                    "pattern"
                ]
            }
        }
    },
    {
        "type": "function",
        "function": {
            "name": "search_alarm",
            "description": "在闹钟中搜索记录。",
            "parameters": {
                "type": "object",
                "properties": {
                    "pattern": {
                        "type": "string",
                        "description": "搜索内容。"
                    },
                    "field": {
                        "type": "string",
                        "description": "指定搜索的字段名。可选字段：'时间'、'重复'、'闹钟名'、'响铃时长（分钟）'、'重复响铃次数'、'响铃间隔时间（分钟）'、'状态'。不指定则搜索所有字段。"
                    },
                    "mode": {
                        "type": "string",
                        "description": "匹配模式。fixed：子串包含匹配（默认）；regex：正则表达式匹配；fuzzy：模糊相似度匹配。",
                        "enum": [
                            "fixed",
                            "regex",
                            "fuzzy"
                        ]
                    },
                    "ignore_case": {
                        "type": "boolean",
                        "description": "是否忽略大小写，默认为true。"
                    },
                    "page": {
                        "type": "integer",
                        "description": "返回第几页结果，每页最多5条。默认为1。"
                    }
                },
                "required": [
                    "pattern"
                ]
            }
        }
    },
    {
        "type": "function",
        "function": {
            "name": "search_contacts",
            "description": "在通讯录中搜索联系人记录。",
            "parameters": {
                "type": "object",
                "properties": {
                    "pattern": {
                        "type": "string",
                        "description": "搜索内容。"
                    },
                    "field": {
                        "type": "string",
                        "description": "指定搜索的字段名。可选字段：'姓名'、'电话号码'、'公司'、'电子邮件'、'住址'、'生日'、'身份证号'、'与本人关系'、'备注'。不指定则搜索所有字段。"
                    },
                    "mode": {
                        "type": "string",
                        "description": "匹配模式。fixed：子串包含匹配（默认）；regex：正则表达式匹配；fuzzy：模糊相似度匹配。",
                        "enum": [
                            "fixed",
                            "regex",
                            "fuzzy"
                        ]
                    },
                    "ignore_case": {
                        "type": "boolean",
                        "description": "是否忽略大小写，默认为true。"
                    },
                    "page": {
                        "type": "integer",
                        "description": "返回第几页结果，每页最多5条。默认为1。"
                    }
                },
                "required": [
                    "pattern"
                ]
            }
        }
    },
    {
        "type": "function",
        "function": {
            "name": "search_meeting",
            "description": "在会议中搜索记录。",
            "parameters": {
                "type": "object",
                "properties": {
                    "pattern": {
                        "type": "string",
                        "description": "搜索内容。"
                    },
                    "field": {
                        "type": "string",
                        "description": "指定搜索的字段名。可选字段：'标题'、'开始日期'、'结束日期'、'开始时间'、'结束时间'、'提醒时间'、'重复'、'参会人'、'会议号'、'地点'、'会议纪要'、'状态'。不指定则搜索所有字段。"
                    },
                    "mode": {
                        "type": "string",
                        "description": "匹配模式。fixed：子串包含匹配（默认）；regex：正则表达式匹配；fuzzy：模糊相似度匹配。",
                        "enum": [
                            "fixed",
                            "regex",
                            "fuzzy"
                        ]
                    },
                    "ignore_case": {
                        "type": "boolean",
                        "description": "是否忽略大小写，默认为true。"
                    },
                    "page": {
                        "type": "integer",
                        "description": "返回第几页结果，每页最多5条。默认为1。"
                    }
                },
                "required": [
                    "pattern"
                ]
            }
        }
    },
    {
        "type": "function",
        "function": {
            "name": "search_notes",
            "description": "在便签中搜索记录。",
            "parameters": {
                "type": "object",
                "properties": {
                    "pattern": {
                        "type": "string",
                        "description": "搜索内容。"
                    },
                    "field": {
                        "type": "string",
                        "description": "指定搜索的字段名。可选字段：'标题'、'内容'、'创建日期'、'创建时间'。不指定则搜索所有字段。"
                    },
                    "mode": {
                        "type": "string",
                        "description": "匹配模式。fixed：子串包含匹配（默认）；regex：正则表达式匹配；fuzzy：模糊相似度匹配。",
                        "enum": [
                            "fixed",
                            "regex",
                            "fuzzy"
                        ]
                    },
                    "ignore_case": {
                        "type": "boolean",
                        "description": "是否忽略大小写，默认为true。"
                    },
                    "page": {
                        "type": "integer",
                        "description": "返回第几页结果，每页最多5条。默认为1。"
                    }
                },
                "required": [
                    "pattern"
                ]
            }
        }
    },
    {
        "type": "function",
        "function": {
            "name": "search_schedule",
            "description": "在日程中搜索记录。",
            "parameters": {
                "type": "object",
                "properties": {
                    "pattern": {
                        "type": "string",
                        "description": "搜索内容。"
                    },
                    "field": {
                        "type": "string",
                        "description": "指定搜索的字段名。可选字段：'标题'、'日期'、'开始时间'、'结束时间'、'提醒时间'、'重复'、'地点'。不指定则搜索所有字段。"
                    },
                    "mode": {
                        "type": "string",
                        "description": "匹配模式。fixed：子串包含匹配（默认）；regex：正则表达式匹配；fuzzy：模糊相似度匹配。",
                        "enum": [
                            "fixed",
                            "regex",
                            "fuzzy"
                        ]
                    },
                    "ignore_case": {
                        "type": "boolean",
                        "description": "是否忽略大小写，默认为true。"
                    },
                    "page": {
                        "type": "integer",
                        "description": "返回第几页结果，每页最多5条。默认为1。"
                    }
                },
                "required": [
                    "pattern"
                ]
            }
        }
    },
    {
        "type": "function",
        "function": {
            "name": "search_sms",
            "description": "在短信中搜索记录。",
            "parameters": {
                "type": "object",
                "properties": {
                    "pattern": {
                        "type": "string",
                        "description": "搜索内容。"
                    },
                    "field": {
                        "type": "string",
                        "description": "指定搜索的字段名。可选字段：'发件人姓名'、'发件人电话号码'、'收件人姓名'、'收件人电话号码'、'日期'、'时间'、'正文'、'附件'、'状态'。不指定则搜索所有字段。"
                    },
                    "mode": {
                        "type": "string",
                        "description": "匹配模式。fixed：子串包含匹配（默认）；regex：正则表达式匹配；fuzzy：模糊相似度匹配。",
                        "enum": [
                            "fixed",
                            "regex",
                            "fuzzy"
                        ]
                    },
                    "ignore_case": {
                        "type": "boolean",
                        "description": "是否忽略大小写，默认为true。"
                    },
                    "page": {
                        "type": "integer",
                        "description": "返回第几页结果，每页最多5条。默认为1。"
                    }
                },
                "required": [
                    "pattern"
                ]
            }
        }
    },
    {
        "type": "function",
        "function": {
            "name": "search_transaction",
            "description": "在交易记录中搜索。",
            "parameters": {
                "type": "object",
                "properties": {
                    "pattern": {
                        "type": "string",
                        "description": "搜索内容。"
                    },
                    "field": {
                        "type": "string",
                        "description": "指定搜索的字段名。可选字段：'项目'、'金额'、'日期'、'时间'、'付款人'、'付款账号'、'收款人'、'收款账号'、'备注'。不指定则搜索所有字段。"
                    },
                    "mode": {
                        "type": "string",
                        "description": "匹配模式。fixed：子串包含匹配（默认）；regex：正则表达式匹配；fuzzy：模糊相似度匹配。",
                        "enum": [
                            "fixed",
                            "regex",
                            "fuzzy"
                        ]
                    },
                    "ignore_case": {
                        "type": "boolean",
                        "description": "是否忽略大小写，默认为true。"
                    },
                    "page": {
                        "type": "integer",
                        "description": "返回第几页结果，每页最多5条。默认为1。"
                    }
                },
                "required": [
                    "pattern"
                ]
            }
        }
    },
    {
        "type": "function",
        "function": {
            "name": "search_transport",
            "description": "在交通记录中搜索。",
            "parameters": {
                "type": "object",
                "properties": {
                    "pattern": {
                        "type": "string",
                        "description": "搜索内容。"
                    },
                    "field": {
                        "type": "string",
                        "description": "指定搜索的字段名。可选字段：'日期'、'出发时间'、'（预计）到达时间'、'出发地'、'目的地'、'交通方式'、'联系人'、'联系方式'、'（预计）费用'、'状态'。不指定则搜索所有字段。"
                    },
                    "mode": {
                        "type": "string",
                        "description": "匹配模式。fixed：子串包含匹配（默认）；regex：正则表达式匹配；fuzzy：模糊相似度匹配。",
                        "enum": [
                            "fixed",
                            "regex",
                            "fuzzy"
                        ]
                    },
                    "ignore_case": {
                        "type": "boolean",
                        "description": "是否忽略大小写，默认为true。"
                    },
                    "page": {
                        "type": "integer",
                        "description": "返回第几页结果，每页最多5条。默认为1。"
                    }
                },
                "required": [
                    "pattern"
                ]
            }
        }
    },
    {
        "type": "function",
        "function": {
            "name": "search_voicemail",
            "description": "在语音留言中搜索记录。",
            "parameters": {
                "type": "object",
                "properties": {
                    "pattern": {
                        "type": "string",
                        "description": "搜索内容。"
                    },
                    "field": {
                        "type": "string",
                        "description": "指定搜索的字段名。可选字段：'来电姓名'、'来电号码'、'日期'、'时间'、'留言内容'。不指定则搜索所有字段。"
                    },
                    "mode": {
                        "type": "string",
                        "description": "匹配模式。fixed：子串包含匹配（默认）；regex：正则表达式匹配；fuzzy：模糊相似度匹配。",
                        "enum": [
                            "fixed",
                            "regex",
                            "fuzzy"
                        ]
                    },
                    "ignore_case": {
                        "type": "boolean",
                        "description": "是否忽略大小写，默认为true。"
                    },
                    "page": {
                        "type": "integer",
                        "description": "返回第几页结果，每页最多5条。默认为1。"
                    }
                },
                "required": [
                    "pattern"
                ]
            }
        }
    },
    {
        "type": "function",
        "function": {
            "name": "search_phone",
            "description": "全局搜索手机中所有应用的记录。使用子串包含匹配，搜索所有字段。可指定field缩小搜索范围。每页返回5条结果，可通过page参数翻页。",
            "parameters": {
                "type": "object",
                "properties": {
                    "pattern": {
                        "type": "string",
                        "description": "搜索内容。"
                    },
                    "field": {
                        "type": "string",
                        "description": "指定搜索的字段名（可选）。不指定则搜索所有字段。不同应用的可用字段不同，常见字段包括：'姓名'、'电话号码'、'与本人关系'、'备注'、'正文'、'留言内容'、'标题'、'内容'等。"
                    },
                    "ignore_case": {
                        "type": "boolean",
                        "description": "是否忽略大小写，默认为true。"
                    },
                    "page": {
                        "type": "integer",
                        "description": "返回第几页结果，每页最多5条。默认为1。"
                    }
                },
                "required": [
                    "pattern"
                ]
            }
        }
    }
]
\end{Verbatim}
\end{tcolorbox}

\captionof{figure}{Specification documents for all retrieval tools.}
\label{fig:retrieval_tools}
\end{center}

\begin{center}
\begin{tcolorbox}[
    colback=white,
    colframe=gray!70!black,
    title=Specification Documents for all Execution Tools,
    coltitle=white,
    fonttitle=\bfseries,
    center title,
    rounded corners,
    boxrule=0.6mm,
    width=\linewidth,
    breakable,
    enhanced,
    left=6pt,
    right=6pt,
    top=4pt,
    bottom=4pt
]

\begin{Verbatim}[breaklines]
[
    {
        "type": "function",
        "function": {
            "name": "Accommodation_Create",
            "description": "创建一个住宿信息。",
            "parameters": {
                "type": "object",
                "properties": {
                    "dateCheckIn": {
                        "type": "string",
                        "description": "入住日期，日期格式为YYYY-MM-DD（例如：'2025-05-20'表示2025年5月20日）。"
                    },
                    "dateCheckOut": {
                        "type": "string",
                        "description": "退宿日期，日期格式为YYYY-MM-DD（例如：'2025-05-20'表示2025年5月20日）。"
                    },
                    "location": {
                        "type": "string",
                        "description": "住宿地点。"
                    },
                    "roomType": {
                        "type": "string",
                        "description": "房间类型。"
                    },
                    "num": {
                        "type": "integer",
                        "description": "房间数量，默认为1。"
                    },
                    "people": {
                        "type": "array",
                        "description": "入住人。",
                        "items": {
                            "type": "string",
                            "description": "入住人姓名。"
                        }
                    },
                    "telephone": {
                        "type": "string",
                        "description": "联系电话。"
                    }
                },
                "required": [
                    "dateCheckIn",
                    "dateCheckOut",
                    "location",
                    "people",
                    "telephone"
                ]
            }
        }
    },
    {
        "type": "function",
        "function": {
            "name": "Clock_CreateAlarm",
            "description": "创建闹钟，支持设置闹钟的触发时间、重复规则等属性",
            "parameters": {
                "type": "object",
                "properties": {
                    "time": {
                        "type": "string",
                        "description": "闹钟触发的具体时间（24小时内的时间，不包含日期），格式为HH:MM:SS（例如：'14:30:00'表示下午2点30分）。"
                    },
                    "repeatDayOfWeek": {
                        "type": "array",
                        "description": "重复规则，类型为数组，用于指定闹钟在哪天重复触发。数字1到7分别表示每个星期一到星期日重复，数字0表示不重复。默认不重复。",
                        "items": {
                            "type": "integer",
                            "description": "重复规则。数字1到7分别表示星期一到星期日，数字0表示不重复。",
                            "enum": [
                                1,
                                2,
                                3,
                                4,
                                5,
                                6,
                                7,
                                0
                            ]
                        }
                    },
                    "name": {
                        "type": "string",
                        "description": "闹钟名称，默认为“闹钟”。"
                    },
                    "alarmDuration": {
                        "type": "integer",
                        "description": "响铃时长（分钟），可以是1分钟，3分钟，5分钟，10分钟。默认为3分钟。",
                        "enum": [
                            1,
                            3,
                            5,
                            10
                        ]
                    },
                    "alarmCount": {
                        "type": "integer",
                        "description": "重复响铃次数，可以是1次，3次，5次，10次。默认为1次。",
                        "enum": [
                            1,
                            3,
                            5,
                            10
                        ]
                    },
                    "alarmInterval": {
                        "type": "integer",
                        "description": "响铃间隔（分钟），可以是5-30分钟。默认为5分钟。",
                        "enum": [
                            5,
                            10,
                            15,
                            20,
                            25,
                            30
                        ]
                    }
                },
                "required": [
                    "time"
                ]
            }
        }
    },
    {
        "type": "function",
        "function": {
            "name": "Contacts_Create",
            "description": "新建一个联系人，并保存到手机通讯录中。",
            "parameters": {
                "type": "object",
                "properties": {
                    "name": {
                        "type": "string",
                        "description": "联系人的姓名。"
                    },
                    "telephone": {
                        "type": "string",
                        "description": "联系人的电话号码。"
                    },
                    "company": {
                        "type": "string",
                        "description": "联系人的公司名称。"
                    },
                    "email": {
                        "type": "string",
                        "description": "联系人的电子邮件地址。"
                    },
                    "address": {
                        "type": "string",
                        "description": "联系人的地址。"
                    },
                    "birthday": {
                        "type": "string",
                        "description": "联系人的生日。"
                    },
                    "identifyCard": {
                        "type": "string",
                        "description": "联系人的身份证号码。"
                    },
                    "relationship": {
                        "type": "string",
                        "description": "联系人与本人的关系。"
                    },
                    "note": {
                        "type": "string",
                        "description": "为联系人添加的备注信息。"
                    }
                },
                "required": [
                    "name",
                    "telephone"
                ]
            }
        }
    },
    {
        "type": "function",
        "function": {
            "name": "Meeting_Create",
            "description": "预订会议，支持设置时间、地点、参与人、会议主题等信息",
            "parameters": {
                "type": "object",
                "properties": {
                    "dateStart": {
                        "type": "string",
                        "description": "会议的开始日期，日期格式为YYYY-MM-DD（例如：'2025-05-20'表示2025年5月20日）"
                    },
                    "dateEnd": {
                        "type": "string",
                        "description": "会议的结束日期，日期格式为YYYY-MM-DD（例如：'2025-05-20'表示2025年5月20日）"
                    },
                    "timeStart": {
                        "type": "string",
                        "description": "会议的开始时间，时间格式为HH:MM:SS（例如：'14:30:00'表示下午2点30分）"
                    },
                    "timeEnd": {
                        "type": "string",
                        "description": "会议的结束时间，时间格式为HH:MM:SS（例如：'14:30:00'表示下午2点30分）"
                    },
                    "timeRemind": {
                        "type": "string",
                        "description": "会议提醒的时间。默认为10分钟前。",
                        "enum": [
                            "准时",
                            "10分钟前",
                            "30分钟前",
                            "1小时前",
                            "1天前",
                            "1周前"
                        ]
                    },
                    "title": {
                        "type": "string",
                        "description": "会议的主题。默认为“无主题”。"
                    },
                    "repeat": {
                        "type": "string",
                        "description": "会议重复规则。默认为不重复。",
                        "enum": [
                            "不重复",
                            "每天",
                            "每周",
                            "每两周"
                        ]
                    },
                    "location": {
                        "type": "array",
                        "description": "会议的地点。",
                        "items": {
                            "type": "string",
                            "description": "地点名称。默认为“无”。"
                        }
                    },
                    "participants": {
                        "type": "array",
                        "description": "会议的参与者，默认已包含自己，只需要填写其他仍需加入的用户即可。",
                        "items": {
                            "type": "string",
                            "description": "会议除自己外的参与者姓名，可以是昵称。"
                        }
                    }
                },
                "required": [
                    "dateStart",
                    "dateEnd",
                    "timeStart",
                    "timeEnd"
                ]
            }
        }
    },
    {
        "type": "function",
        "function": {
            "name": "TodoList_Create",
            "description": "新建一个待办事项。",
            "parameters": {
                "type": "object",
                "properties": {
                    "content": {
                        "type": "string",
                        "description": "待办事项的内容。"
                    },
                    "date": {
                        "type": "string",
                        "description": "待办事项的具体日期，日期格式为YYYY-MM-DD（例如：'2025-05-20'表示2025年5月20日）"
                    },
                    "time": {
                        "type": "string",
                        "description": "待办事项的开始时间，时间格式为HH:MM（例如：'14:30'表示下午2点30分）"
                    }
                },
                "required": [
                    "content",
                    "date",
                    "time"
                ]
            }
        }
    },
    {
        "type": "function",
        "function": {
            "name": "Schedule_Create",
            "description": "新建一个日程。",
            "parameters": {
                "type": "object",
                "properties": {
                    "date": {
                        "type": "string",
                        "description": "日程触发的具体日期，日期格式为YYYY-MM-DD（例如：'2025-05-20'表示2025年5月20日）"
                    },
                    "title": {
                        "type": "string",
                        "description": "日程的标题，用于标识日程的主题。默认为“无主题”。"
                    },
                    "timeStart": {
                        "type": "string",
                        "description": "日程开始的具体时间，时间格式为HH:MM:SS（例如：'14:30:00'表示下午2点30分）"
                    },
                    "timeEnd": {
                        "type": "string",
                        "description": "日程结束的具体时间，时间格式为HH:MM:SS（例如：'14:30:00'表示下午2点30分）"
                    },
                    "timeRemind": {
                        "type": "string",
                        "description": "日程提醒的时间。默认为10分钟前。",
                        "enum": [
                            "准时",
                            "10分钟前",
                            "30分钟前",
                            "1小时前",
                            "1天前",
                            "1周前"
                        ]
                    },
                    "repeat": {
                        "type": "string",
                        "description": "日程重复规则。默认为不重复。",
                        "enum": [
                            "不重复",
                            "每天",
                            "每周",
                            "每月",
                            "每年"
                        ]
                    },
                    "location": {
                        "type": "string",
                        "description": "日程的地点。默认为“无”。"
                    }
                },
                "required": [
                    "date",
                    "timeStart",
                    "timeEnd"
                ]
            }
        }
    },
    {
        "type": "function",
        "function": {
            "name": "Messages_Send",
            "description": "向某个电话号码发送短信，包括正文和附件。",
            "parameters": {
                "type": "object",
                "properties": {
                    "telephone": {
                        "type": "string",
                        "description": "需要接受短信的电话号码。"
                    },
                    "text": {
                        "type": "string",
                        "description": "需要发送的短信正文，仅支持文本信息。"
                    },
                    "attach": {
                        "type": "array",
                        "description": "需要发送的短信附件，通过指定文件路径发送，最多支持同时发送10个附件。",
                        "items": {
                            "type": "string",
                            "description": "文件路径。"
                        }
                    }
                },
                "required": [
                    "telephone",
                    "text"
                ]
            }
        }
    },
    {
        "type": "function",
        "function": {
            "name": "Consumption_Create",
            "description": "向他人付款。",
            "parameters": {
                "type": "object",
                "properties": {
                    "item": {
                        "type": "string",
                        "description": "付款条目，如转账、购物等。"
                    },
                    "amount": {
                        "type": "number",
                        "description": "交易金额，精确到小数点后两位。"
                    },
                    "receivingName": {
                        "type": "string",
                        "description": "收款人。"
                    },
                    "receivingAccount": {
                        "type": "string",
                        "description": "收款账户。"
                    },
                    "note": {
                        "type": "string",
                        "description": "备注。"
                    }
                },
                "required": [
                    "item",
                    "amount",
                    "receivingName",
                    "receivingAccount"
                ]
            }
        }
    },
    {
        "type": "function",
        "function": {
            "name": "Transport_Create",
            "description": "创建一个交通信息。",
            "parameters": {
                "type": "object",
                "properties": {
                    "date": {
                        "type": "string",
                        "description": "出发日期，日期格式为YYYY-MM-DD（例如：'2025-05-20'表示2025年5月20日）。202"
                    },
                    "time": {
                        "type": "string",
                        "description": "出发时间，时间格式为HH:MM:SS（例如：'14:30:00'表示下午2点30分）。如果是现在出发，可以输入'now'。"
                    },
                    "start": {
                        "type": "string",
                        "description": "出发地。打车时如果是当前位置，可以输入'current'。"
                    },
                    "end": {
                        "type": "string",
                        "description": "目的地。"
                    },
                    "passenger": {
                        "type": "array",
                        "description": "乘车人。",
                        "items": {
                            "type": "string",
                            "description": "乘车人姓名。"
                        }
                    },
                    "telephone": {
                        "type": "string",
                        "description": "乘车人联系电话。"
                    },
                    "transportType": {
                        "type": "string",
                        "description": "交通方式。",
                        "enum": [
                            "汽车",
                            "火车",
                            "飞机"
                        ]
                    }
                },
                "required": [
                    "date",
                    "time",
                    "start",
                    "end",
                    "passenger",
                    "telephone",
                    "transportType"
                ]
            }
        }
    },
    {
        "type": "function",
        "function": {
            "name": "Phone_Call",
            "description": "给某个电话号码打电话，支持语音电话和视频电话两种类型。",
            "parameters": {
                "type": "object",
                "properties": {
                    "telephone": {
                        "type": "string",
                        "description": "需要拨打的电话号码。"
                    },
                    "type": {
                        "type": "string",
                        "description": "电话类型，枚举值：- VOICE 语音电话 - VIDEO 视频电话，默认为VOICE",
                        "enum": [
                            "VOICE",
                            "VIDEO"
                        ]
                    }
                },
                "required": [
                    "telephone"
                ]
            }
        }
    }
]
\end{Verbatim}
\end{tcolorbox}

\captionof{figure}{Specification documents for all execution tools.}
\label{fig:execution_tools}
\end{center}

\clearpage

\section{Comparison with Existing Benchmarks}
\label{sec:appendix_comparison}

As described in Section~\ref{sec:data_analysis}, \spieval{} exhibits five key characteristics, including diverse scenarios, challenging tasks, scattered information, controllable environments, and verifiable outcomes. To further illustrate its strengths, we compare \spieval{} with existing benchmarks, with the results summarized in Table~\ref{tab:comparison}.

\begin{table}[h]
\centering
\caption{Comparison of \spieval{} with existing benchmarks. For dialogue- and document-based benchmarks, \#Tasks reports the number of question--answer instances and \#Apps is not applicable. \cmark{} and \xmark{} indicate the presence and absence of a systematic benchmark-level property, respectively.}
\label{tab:comparison}
\resizebox{\linewidth}{!}{%
\begin{tabular}{lrrcccccc}
\toprule
\multirow{2}{*}{\textbf{Benchmark}}
& \multicolumn{2}{c}{\textbf{Basic Information}}
& \multicolumn{4}{c}{\textbf{Task Setting}}
& \multicolumn{2}{c}{\textbf{Benchmark Design}} \\
\cmidrule(lr){2-3}\cmidrule(lr){4-7}\cmidrule(lr){8-9}
& \textbf{\#Tasks}
& \textbf{\#Apps}
& \textbf{Underspecified}
& \textbf{\shortstack{Scattered Personal\\Information}}
& \textbf{\shortstack{Proactive\\Retrieval}}
& \textbf{\shortstack{Action\\Execution}}
& \textbf{Verifiable}
& \textbf{\shortstack{Human-\\Curated}} \\
\midrule
AppWorld & 750 & 9 & \xmark & \xmark & \cmark & \cmark & \cmark & \xmark \\
Gaia2 & 1,120 & 12 & \xmark & \xmark & \cmark & \cmark & \cmark & \xmark \\
SAPA-Bench & 7,138 & 50 & \xmark & \xmark & \xmark & \xmark & \xmark & \xmark \\
HiCUPID & 60,000 & -- & \xmark & \xmark & \xmark & \xmark & \xmark & \xmark \\
PersonaBench & 582 & -- & \xmark & \xmark & \xmark & \xmark & \cmark & \xmark \\
\midrule
\spieval{} & 250 & 10 & \cmark & \cmark & \cmark & \cmark & \cmark & \cmark \\
\bottomrule
\end{tabular}%
}
\end{table}

\end{CJK}

\end{document}